\documentclass[journal]{IEEEtran}

\usepackage{cite}
\usepackage{amsmath,amssymb,amsfonts}
\usepackage{algorithmic}
\usepackage{graphicx}
\usepackage{textcomp}
\usepackage[utf8]{inputenc} 
\usepackage[T1]{fontenc}    
\usepackage{hyperref}       
\usepackage{url}            
\usepackage{booktabs}       
\usepackage{amsfonts}       
\usepackage{amsmath}
\usepackage{nicefrac}       
\usepackage{microtype}      
\usepackage{xcolor}         
\usepackage{graphicx}
\usepackage{wrapfig}
\usepackage{pifont}
\usepackage{multirow}
\usepackage{xcolor}
\usepackage{textcomp}
\usepackage{color,soul}

\def\textbf#1{{\bf #1}}

\def\BibTeX{{\rm B\kern-.05em{\sc i\kern-.025em b}\kern-.08em
    T\kern-.1667em\lower.7ex\hbox{E}\kern-.125emX}}
\begin{document}

\title{Generative Semantic Communication:\\Diffusion Models Beyond Bit Recovery}
\author{Eleonora Grassucci, Sergio Barbarossa, \IEEEmembership{Fellow, IEEE} and Danilo Comminiello,\thanks{This work was supported by the European Union under the Italian National Recovery and Resilience Plan (PNRR) of NextGenerationEU, partnership on
``Telecommunications of the Future” (PE00000001 - program RESTART), and by Huawei Technology France SASU, under agreement N.TC20220919044.} \IEEEmembership{Senior Member, IEEE}\\
Dept. of Information Engineering, Electronics, and Telecommunication, Sapienza University of Rome, Rome, Italy\\Corresponding author's email: eleonora.grassucci@uniroma1.it.}

\maketitle

\begin{abstract}
Semantic communications have the potential to play a key role in next-generation AI-native communication systems, especially when combined with the expressivity power of generative models.
In this paper, we focus on image transmission, and we present a novel generative-based semantic communication framework, whose core is a generative model operating at the receiver side. This model regenerates images suitable for downstream applications such as detection, reconstruction, and positioning of semantically relevant objects in the scene observed from the sensors present at the sender side. We devise the encoding rule to transmit only what is strictly relevant to trigger the generative model to fulfill the scope of the transmission. Furthermore, we propose a training strategy to make the generative model robust against additive noise due to propagation through the communication channel. 
We prove, through an in-depth assessment of multiple scenarios, that our method outperforms existing solutions in generating high-quality images with preserved semantic information even in cases where the received conditioning content is significantly degraded or compressed. More specifically, our results show that objects, locations, and depths are still recognizable even in the presence of highly noisy conditions of the communication channel or at very low bits per pixel.
\end{abstract}

\begin{IEEEkeywords}
Generative Semantic Communication, Diffusion Models, Semantic Communication, Image Transmission
\end{IEEEkeywords}

\maketitle

\section{Introduction}
\label{sec:introduction}
The upcoming sixth generation (6G) of wireless networks is expected to bring a radical change in the design and development of communication systems by leveraging advanced Artificial Intelligence (AI) tools \cite{Zhang2024CommSurvey, You2024NextGA, Luo2022WC}. In particular, the emerging paradigm of semantic communications can greatly benefit from effective knowledge representation tools made available from AI algorithms. Semantic communications focus on using context-aware representations that are most suitable for the communication task. As an example of such an application, let us suppose that a vehicle needs to send the images or videos captured by its onboard cameras to a roadside unit (RSU) for processing. In this application, it is crucial to make quick and accurate decisions about potential dangers. Hence, it is fundamental to tune the encoding strategy to fulfill the above goal, without necessarily requiring the receiver to exactly reproduce the images or videos as they are taken from the vehicle's sensors. What really matters in such an application is to enable the receiver to identify relevant objects in the scene, like pedestrians or cars, and to position them accurately, possibly in a 3D environment. From this perspective, the coding strategy can take advantage of intelligent tools, such as the extraction of image segmentation maps, image generation, 3D depth estimation from 2D images, etc. This approach can greatly benefit from deep neural networks (DNN), properly trained on data sets built for this purpose.  


In this context, deep generative models can play a key role due to their ability to generate multimedia content from highly compressed information. Building on these tools, the communication paradigm becomes how to select the information to be transmitted to trigger the generative model used at the receiver side to produce a representation conditioned to what has been transmitted and suitable for the application that is going to run on that representation.
Among the variety of generative models, denoising diffusion probabilistic models (DDPMs) \cite{Ho2020DDPM} have been shown to exhibit remarkable achievements in a plethora of real-world generation tasks \cite{saharia2022photorealistic, Rombach2021latent, Ghosal2023TexttoAudioGU, hong2023cogvideo}.
Among such significant results, diffusion models are able to produce photorealistic images preserving the semantic layout \cite{Wang2022SemanticIS, Xue2023FreestyleLS} in the so-called semantic image synthesis (SIS) task. The success of these models in countless domains, and especially in the SIS task, inspired us to involve them in semantic communications. However, typically such models have been studied on clean data and their extension to data altered by compression or propagation through a noisy channel has not been investigated yet. This issue may prevent the effective deployment of diffusion models for communication purposes.

In this paper, we make a step towards bridging semantic communications and state-of-the-art generative models by presenting a novel generative semantic communication framework that is robust against distortion due to propagation through a noisy channel. The core of our framework is a robust semantic diffusion model that generates photorealistic images preserving the shapes and relative positioning of semantically relevant objects present in the observed scene and enabling 3D depth estimation at the receiver side. The sender extracts and transmits a semantic map, which is a compact representation of the observed scene relevant to the application. The receiver collects a corrupted version of the transmitted data and applies fast denoising to the map before using it as a conditioning for the generative process. The whole framework is made robust to propagation through bad channel conditions, ensuring that even in the case of extremely degraded received information, the semantically relevant objects, their positions, and their depths are still recognizable in the synthesized images, differently from existing approaches or large-scale pretrained generative models. To assess the performance of the proposed approach, we introduce performance metrics that are relevant to the application at hand and show how to achieve a better trade-off between compression rate and these ad hoc performance metrics, with respect to state-of-the-art generative models. 
We test our approach under different channel conditions and datasets and we show how our method can generate photorealistic images consistent with the transmitted semantic information even in the case of extremely corrupted received layouts and in cases where the transmitted data rate is very low. 
The main contributions of this work can be summarized as follows:

\begin{itemize}
\item We propose an image transmission system that is compliant with legacy systems and allows the recovery of semantic information about a class of relevant objects even transmitting at a very low coding rate, thanks to the use of generative models at the receiver side;
    \item We introduce a novel noise-robust semantic diffusion model capable of regenerating high-quality images even in the case of highly deteriorated received data due to channel noise and large compression rates.
    \item We show how the proposed method can be effectively involved for different downstream tasks, such as object detection and depth estimation, with an efficient usage of resources, such as bandwidth.
\end{itemize}

The remainder of the paper is organized as follows. Section~\ref{sec:works} reports related works, Section~\ref{sec:method} presents the proposed framework and the problem setting, while Section~\ref{sec:exp} shows the experimental evaluation and the analysis of the results. Finally, conclusions are drawn in Section~\ref{sec:conc}.


\section{Related Work}
\label{sec:works}
Semantic communication is expected to play a key role in 6G networks \cite{Zhang2024CommSurvey, You2024NextGA, Strinati20206GNB, Luo2022WC, Huang2023JointTA, Qin2021SemanticCP}. The use of deep neural networks (DNNs) as a way to perform joint source-channel coding (JSCC) in wireless communications has been thoroughly investigated \cite{xu2023deep}. 
Leveraging DNNs as a way to capture semantic aspects of what is being coded has attracted significant research efforts in the last years, influencing several applications ranging from image compression \cite{Patwa2020SemanticPreservingIC, Wang2019AnED} and image forgery detection \cite{revisorecojione1} to video compression/transmission \cite{Jiang2022WirelessSC, Shakarji2019AIPR} and compressed deepfake video dection \cite{revisorecojione1}, and it is expected to increase its impact in many more fields in the next years \cite{Dai2021CommunicationBT}.

Later, a new branch of artificial intelligence gained attention, relying on the capabilities of DNNs to generate new content. In such a generative modeling branch, diffusion models represent a real breakthrough, showing impressive results in several generation tasks, ranging from image \cite{Nichol2021GLIDETP, saharia2022photorealistic, Rombach2021latent, You2023CoBITAC} to audio \cite{Ghosal2023TexttoAudioGU, Popov2023OT, Huang2023MakeAnAudioTG, turetzky22_interspeech} or video generation \cite{hong2023cogvideo, Singer2022MakeAVideoTG, Gu2023SeerLI, Jiang2023Text2PerformerTH}. Diffusion models synthesize samples starting from a standard Gaussian distribution and by performing an iterative denoising process up to the desired new content. This process makes diffusion model generation far stable than generative adversarial networks \cite{Croitoru2023TPAMI}. Among the tasks in which diffusion models stand out, there is semantic image synthesis (SIS), which consists in generating images coherent with a given semantic layout. Although most SIS approaches are based on generative adversarial networks \cite{tan2021diverse, park2019SPADE, schonfeld2021you, Zhu2020SemanticallyMI, liu2019learning}, in the last year, a novel SIS model outperforms other approaches by involving a diffusion model to synthesize semantically consistent high-quality scenes \cite{Wang2022SemanticIS}.

More recently, generative semantic communication methods have been introduced \cite{Barbarossa2023COMMAG, Grassucci2024generativeai}. Among them, generative adversarial networks have been the first generative tool to be involved in tasks such as image compression or denoising \cite{Han2022GenHighEff, Gunduz2022GenSem}. Overall, previous generative communication frameworks have often been limited to rather simple models such as small VAEs \cite{Malur2020VAE, Estiri2020AVA} or pretrained GAN generators \cite{Gunduz2022GenSem}. In addition, normalizing flows have started to be involved in semantic communications to increase framework expressiveness \cite{Han2022GenerativeMB}. 
A combination of GAN-based generators with vector quantization in order to find out a suitable trade-off between image recovery and preservation of semantic content, even when transmitting at very low coding rates, was recently proposed \cite{pezone2025sq}.
In the last year, the interest in using large generative \cite{Grassucci2024ICASSPOv, Grassucci2024generativeai, Yang2024Arxiv, Guo2024TWC, Wu2025TWC} and foundation \cite{Jiang2024WCL} models for semantic communication has grown, as testified by recent works \cite{You2024arxiv, Chen2024Network, Liang2024TCCN, Cheng2023AWA, Qiao2024LatencyAwareGS, Xia2023GenerativeAF}.

\section{Proposed Method}
\label{sec:method}

In this paper, we present a novel generative semantic communication framework based on denoising diffusion probabilistic models (DDPMs) for synthesizing high-quality images in order to preserve the transmitted semantic information for the application underlying the communication. The proposed framework is compliant with the legacy system as it allows the incorporation of state-of-the-art source and channel coding. The key idea is to transmit one-hot encoded semantic maps of the images, which are suitable for strong compression, and use DDPMs at the receiver side to regenerate images that are semantically equivalent to the transmitted ones and enable an effective implementation of downstream tasks such as object detection and 3D depth estimation.
\begin{figure*}
    \centering
    \includegraphics[width=0.8\linewidth]{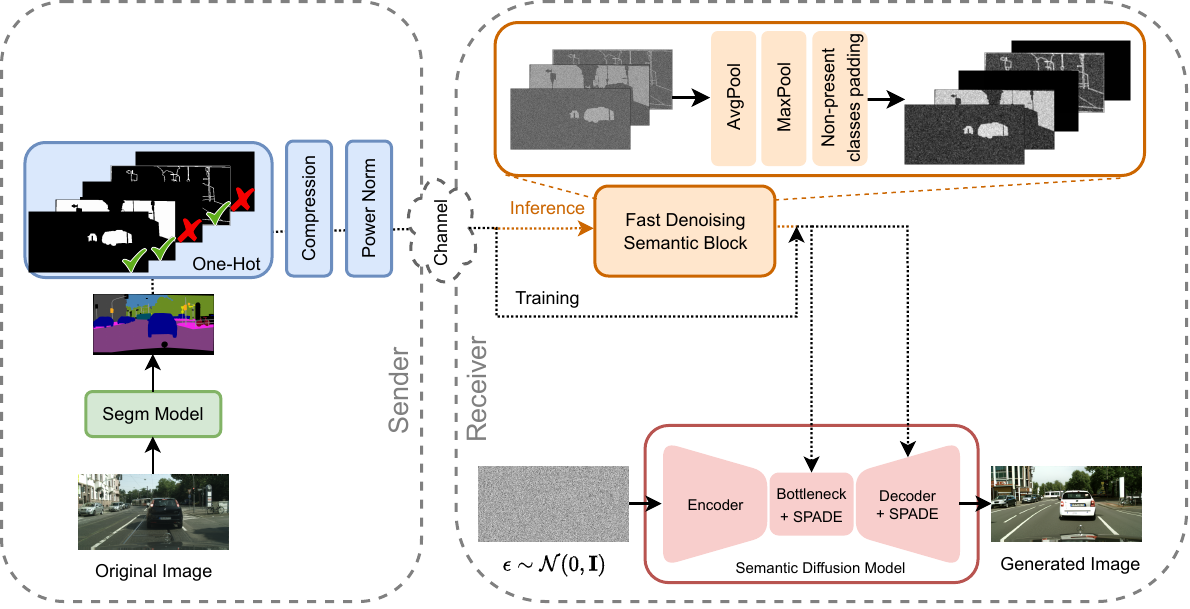}
    \caption{Proposed generative semantic communication framework. The sender transmits one-hot, compressed, and normalized encoded maps over the noisy channel. The receiver takes the noisy maps and directly involves them to train the semantic diffusion model. During inference, the receiver applies fast denoising to the semantic information in order to improve image quality.}
    \label{fig:arch}
\end{figure*}

\subsection{Generative Semantic Communication Framework}
We consider, as an example of application, a moving vehicle that takes pictures and sends them to a roadside unit (RSU), whose goal is to process the received data to correctly detect, identify, and position relevant objects present in the scene observed by the vehicle, in a 3D environment. 
From this point of view, it is not necessary to enable the receiver to recover an image that looks exactly like the one collected onboard the car, but only to enable the RSU to make a decision as accurately as possible in the shortest possible time. Therefore, the idea is to transmit only what is strictly necessary to achieve such a goal most effectively, taking the wireless channel degradation into account. 
Figure \ref{fig:arch} presents the proposed architecture, including both the sender and receiver sides.

\textbf{Sender.}
The first step on the transmitter side is the extraction of the semantic map from the original image, using standard segmentation models. In the autonomous driving scenario, there are already algorithms able to extract the semantic map in real time, see, e.g. \cite{elhassan2024real}. In particular, Tesla has already deployed the technology to analyze raw images and perform semantic segmentation and object detection in their per-camera networks in real time, as reported in \cite{teslaAI}.
The semantic map is then compressed before transmission. We propose to encode the semantic map using a one-hot binary encoder. The output of the encoder is then compressed using state-of-the-art image compression, e.g., Better Portable Graphics (BPG). Since the one-hot encoded map is binary and constant within each object, the combination of binary one-hot encoding and BPG produces a highly compressed file that retains all valuable semantic information. As we will see in the numerical section, the transmission of this file is very robust against degradations due to additive noise, even in the presence of significant noise. 
\textbf{Receiver.}
The receiver aims to accomplish the scope of the communication, whether it is object recognition or depth estimation, among others. To successfully accomplish the goal, taking also into account channel degradation, the first step of the receiver is to regenerate an image that is semantically equivalent to the transmitted one, using the received data. In our example of application, semantic equivalence means that the reconstructed image should contain all relevant objects present in the original image, with the correct shape and relative positioning between each other. This step is necessary; otherwise, tasks like depth estimation would not perform well, if applied directly to the semantic maps, especially when this information is corrupted by the channel noise.

To correctly regenerate the received semantic content, we present a novel Denoising Diffusion Probabilistic Model (DDPM), whose generation process is guided by the received semantics. The model starts from a random set of independent Gaussian random variables, with zero mean and unit variance, i.e. $\mathbf{x}_0 \sim \mathcal{N}(\mathbf{0}, \mathbf{I})$, and progressively removes the noise, while being conditioned by the segmentation maps, as shown in Fig. \ref{fig:arch}. To improve the quality of the generated images during inference, we propose a fast-denoising semantic (FDS) block, whose scope is to attenuate the impact of random perturbations of the received map resulting from propagation through a noisy channel. Furthermore, since the received segmentation map is typically affected by noise, we train the diffusion model with noisy maps and let the network weights adapt to different channel conditions. It is important to note that, differently from previous methods \cite{Shao2021LearningTC, Gunduz2022GenSem}, our receiver does not need to be aware of the channel conditions and it exhibits good results also in the case of adverse channel conditions. The single blocks of our system are detailed in the following sections.

\textbf{Fast Denoising Semantic Block.} The fast-denoising semantic (FDS) block takes in input the corrupted one-hot encoded map $\hat{\mathbf{y}}$ that has been transmitted over the noisy channel. FDS applies a fast denoising taking into account the 0-1 (binary) nature of the maps. In detail, FDS proceeds by concatenating the followinf steps:
\begin{equation}
    \mathbf{y} = \text{Pad}\left(\text{MaxPool}\left(\text{AvgPool}\left(\hat{\mathbf{y}}\right)\right)\right).
\end{equation}
First, the average pooling removes noise spikes in the maps. Then, since the maps comprise large $0/1$ regions, where $1$ corresponds to areas where the class is present and $0$ to empty spaces, the MaxPool operation mainly keeps the $1$s regions only and discards spiky values. Finally, FDS pads the clean missing classes that have been removed on the sender side.

\subsection{Semantic Diffusion Model}

The core of our generative semantic communication framework is the semantic diffusion model that generates images by preserving the transmitted semantic information (i.e., class of the object, position, shape, and dimension).

\textbf{Conditional Diffusion Model.} The diffusion model is composed of a forward diffusion process that starts from an image, represented as a vector $\mathbf{x}_0$, and progressively adds Gaussian noise with zero mean and variance that changes over time according to a schedule $\beta_1, ..., \beta_T$. Setting $\alpha_t := \prod_{s=1}^t (1-\beta_s)$, the forward process, at each time step $t$ is characterized by a conditional pdf:
\begin{equation}
    q(\mathbf{x}_t|\mathbf{x}_0) = \mathcal{N}(\mathbf{x}_t; \sqrt{\alpha_t}\mathbf{x}_0,(1-\alpha_t)\mathbf{I}).
\end{equation}
While injecting noise, the model infers a backward process that, step-by-step, learns how to denoise the image. Given a conditioning vector $\mathbf{y}$, the reverse process is a Markov chain with learned Gaussian transitions that starts at $p(\mathbf{x}_T) \sim \mathcal{N}(0, \mathbf{I})$ and whose pdf can be written as:
\begin{equation}
    p_\theta(\mathbf{x}_{0:T}|\mathbf{y}) = p(\mathbf{x}_T) \prod_{t=1}^Tp_\theta(\mathbf{x}_{t-1}|\mathbf{x}_t, \mathbf{y}),
\end{equation}
with $p_\theta(\mathbf{x}_{t-1}|\mathbf{x}_t, \mathbf{y}) = \mathcal{N}(\mathbf{x}_{t-1}; \mathbf{\mu}_\theta(\mathbf{x}_t, \mathbf{y}, t), \sigma_\theta(\mathbf{x}_t, \mathbf{y}, t))$. 
This reverse diffusion process is grounded in the Markovian structure and Gaussian transition kernels, which ensure that the learned distribution progressively approximates the true data distribution through iterative denoising steps, as established in foundational diffusion model literature \cite{Ho2020DDPM}.


\textbf{Encoder.} The U-Net \cite{Ronneberger2015UNET} encoder comprises an input convolution and a stack of encoder blocks with downsampling. The encoder block interleaves a convolution layer, a Sigmoid Linear Unit (SiLU) activation \cite{Swish2017} taking the form of $\mathbf{y} = \mathbf{x} \cdot  \text{Sigmoid}(\mathbf{x})$, and a group normalization \cite{Wu2018GroupN}.
The latter, given an input tensor $\mathbf{x} \in \mathbb{R}^{N \times C \times H \times W}$, where $N$ is the batch size, $C$ is the number of channels, and $H, W$ are spatial dimensions, divides the $C$ channels into $G$ groups, each containing $C_G = \frac{C}{G}$ channels. Then, for each group $g$ and input feature $x_{i,j,k,l}$ in the group, compute the mean and variance:
\begin{align}
\mu_g &= \frac{1}{|G|} \sum_{\mathbf{x} \in G} \mathbf{x}, \quad
\sigma_g^2 = \frac{1}{|G|} \sum_{\mathbf{x} \in G} (\mathbf{x} - \mu_g)^2,
\end{align}
where $|G| = C_G \cdot H \cdot W$ is the number of elements in the group. Finally, normalize each feature:
\begin{equation}
\hat{\mathbf{x}}_{i,j,k,l} = \frac{\mathbf{x}_{i,j,k,l} - \mu_g}{\sqrt{\sigma_g^2 + \epsilon}},
\end{equation}
where $\epsilon$ is a small constant introduced to ensure numerical stability.

The block also implements a fully-connected layer with weights $\mathbf{W}$ and bias $\mathbf{b}$ to inject the time information $t$ by scaling and shifting the mid-activation $\mathbf{a}$ by $\mathbf{a}_{i+1} = \mathbf{W}(t) \cdot \mathbf{a}_i + \mathbf{b}(t)$. Furthermore, at resolutions $32\times32$, $16\times16$, and $8\times8$ the encoder involves attention modules with skip connections. Given $\mathbf{x}$ input and $\mathbf{y}$ output of the attention block, and four $1\times1$ convolutions with weights $\mathbf{W}_f, \mathbf{W}_g, \mathbf{W}_h,$ and $\mathbf{W}_v$, we define $f(\mathbf{x})=\mathbf{W}_f \mathbf{x}, g(\mathbf{x})=\mathbf{W}_g \mathbf{x}$ and $h(\mathbf{x})=\mathbf{W}_h \mathbf{x}$, arriving at

\begin{equation}
    \mathcal{M}(u,v) = \frac{f(\mathbf{x}_u)^\top g(\mathbf{x}_v)}{\|f(\mathbf{x}_u)\| \|g(\mathbf{x}_v)\|},
\end{equation}

\begin{equation}
    \mathbf{y}_u = \mathbf{x}_u + \mathbf{W}_v \sum_v \text{softmax}_v (\alpha \mathcal{M}(u, v)) \cdot h(\mathbf{x}_v),
\end{equation}

whereby the spatial dimension indexes are $u \in [1, H], v \in [1, W]$.

\begin{figure}
    \centering
    \includegraphics[width=\linewidth]{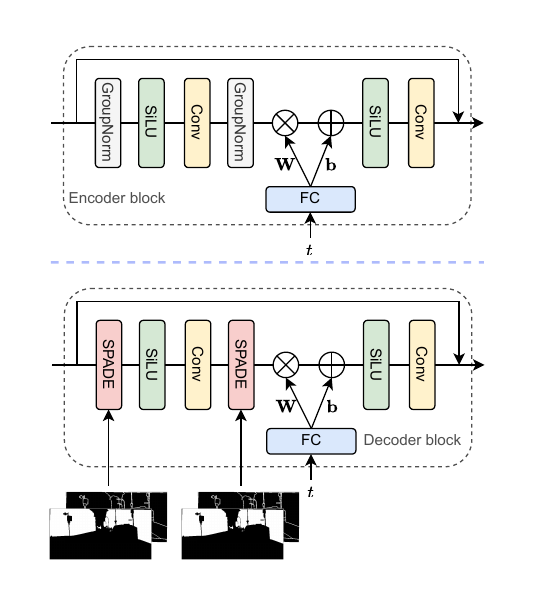}
    \caption{Encoder and decoder blocks of our U-Net-based semantic diffusion model. The SPADE module in the decoder allows the semantic conditioning.}
    \label{fig:blocks}
\end{figure}

\textbf{Decoder.} The decoder blocks are crucial for the semantic conditioning of the whole model. To fully exploit the semantic information, the decoder blocks implement spatially-adaptive normalization (SPADE) \cite{park2019SPADE} that replaces group normalization in the encoder. The SPADE module introduces semantic content in the data flow by adjusting the activations $\mathbf{a}_i$ as follows:

\begin{equation}
    \mathbf{a}_{i+1} = \mathbf{\gamma}_i(\mathbf{x}) \cdot \text{Norm}(\mathbf{a}_i) + \mathbf{b}_i(\mathbf{x}),
\end{equation}
where $\text{Norm}(\cdot)$ indicates group normalization, and $\mathbf{\gamma}_i, \mathbf{b}_i$ are the spatially-adaptive weights and biases learned from the conditioning semantic map with convolutional layers.
To adapt DDPMs to the semantic communication setting, we condition the reverse diffusion process on a compressed semantic map, following the conditional DDPM formulation \cite{SahariaSR3}. This setup ensures that the generative process remains anchored to semantically informative signals even under extreme noise conditions, theoretically supporting the robustness of semantic generation.
At each step, we simulate varying channel conditions by sampling the noise variance in $\{ 0.9, 0.6, 0.36, 0.22, 0.13, 0.05, 0.00 \}$ corresponding to PSNRs in $\{1, 5, 10, 15, 20, 30, 100\}$, weighting perfect channel conditions (PSNR$=100$) higher. We note that incorporating channel noise during training heavily impacts the quality of generated images in inference. A visual representation of the encoder and decoder blocks is shown in Fig. \ref{fig:blocks}.


\subsection{Loss Functions}
\label{subsec:losses}

We train the semantic diffusion model with a combination of two loss functions. Considering an input image $\mathbf{x}$ and the sequence of time steps $t \in \{0, ..., T\}$, the corresponding noisy image $\tilde{\mathbf{x}}$ at time $t$ is built as $\tilde{\mathbf{x}} = \sqrt{\alpha_t}\mathbf{x} + \sqrt{1-\alpha_t}\epsilon$. The noise is sampled from $\epsilon \sim \mathcal{N}(\mathbf{0}, \mathbf{I})$ and  $\alpha_t$ is the noise scheduler at time $t$, where the maximum timestep is $T=1000$. The model tries to predict the noise $\epsilon$ to reconstruct the reference image $\mathbf{x}$ according to the guidance of the semantic map $\mathbf{y}$. The denoising loss function $\mathcal{L}_{\text{d}}$ takes the form of

\begin{equation}
\mathcal{L}_{\text{d}}=\mathbb{E}_{t,\mathbf{x},\epsilon} \left[\left\| \epsilon - \epsilon_\theta \left(\sqrt{\alpha_t} \mathbf{x} + \sqrt{1-\alpha_t}\epsilon, \mathbf{y}, t\right)\right\|_2\right].
\label{eq:recon}
\end{equation}

To improve the generated images log-likelihood, the model is trained to predict variances too, following \cite{Quinn2021Improved}, and employing the KL divergence between the predicted distribution $p_\theta(\mathbf{x}_{t-1} | \mathbf{x}_t, \mathbf{y})$ and the diffusion process posterior $q(\mathbf{x}_{t-1} | \mathbf{x}_t, \mathbf{x}_0)$:

\begin{equation}
    \mathcal{L}_{\text{KL}} = \text{KL}(p_\theta(\mathbf{x}_{t-1} | \mathbf{x}_t, \mathbf{y}) \| q(\mathbf{x}_{t-1} | \mathbf{x}_t, \mathbf{x}_0)). 
\label{eq:kl}
\end{equation}

The resulting loss function is defined as the denoising loss function $\mathcal{L}_{\text{d}}$ plus a penalty given by the KL divergence $\mathcal{L}_{\text{KL}}$, weighted by a nonnegative coefficient $\lambda$ to balance the two terms:

\begin{equation}
    \mathcal{L} = \mathcal{L}_{\text{d}} + \lambda \,\mathcal{L}_{\text{KL}}.
\label{eq:elbo}
\end{equation}

Finally, we note that equations \eqref{eq:recon}, \eqref{eq:kl}, and \eqref{eq:elbo} are the per-timestep components of the evidence-lower-bound (ELBO) for conditional DDPMs \cite{Quinn2021Improved}. Hence, minimizing \eqref{eq:elbo} tightens a variational bound on the conditional data log-likelihood $\log p(\mathbf{x}|\mathbf{y})$.

\begin{figure*}
    \centering
    \includegraphics[width=0.8\linewidth]{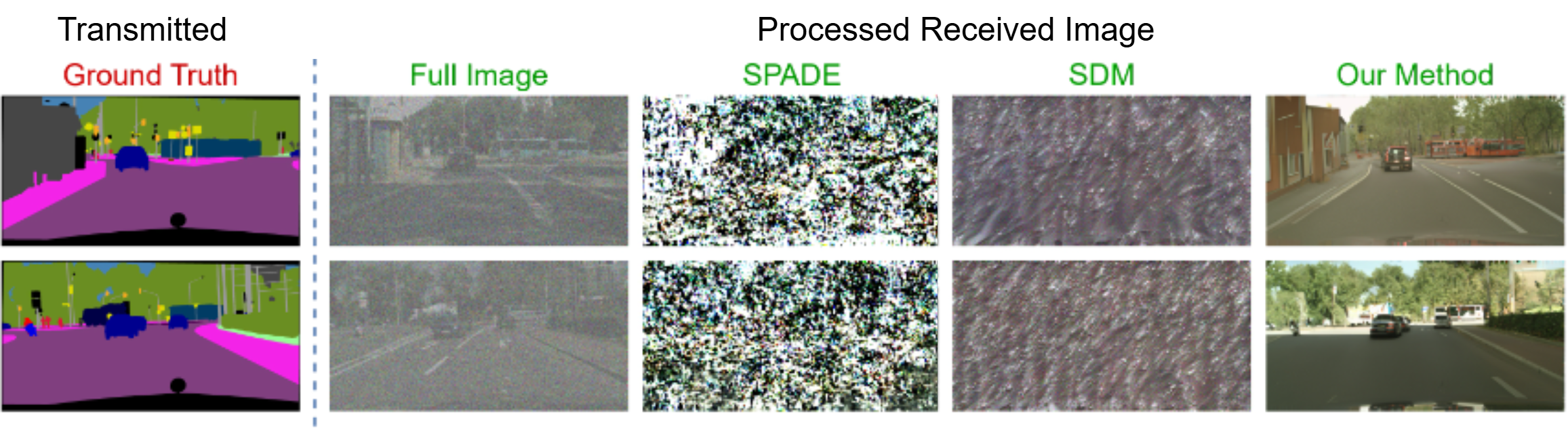}
    \caption{Comparison among the methods for transmitted semantics and a fixed PSNR value of $10$.}
    \label{fig:comp}
\end{figure*}
\begin{figure*}
    \centering
\includegraphics[width=\linewidth]{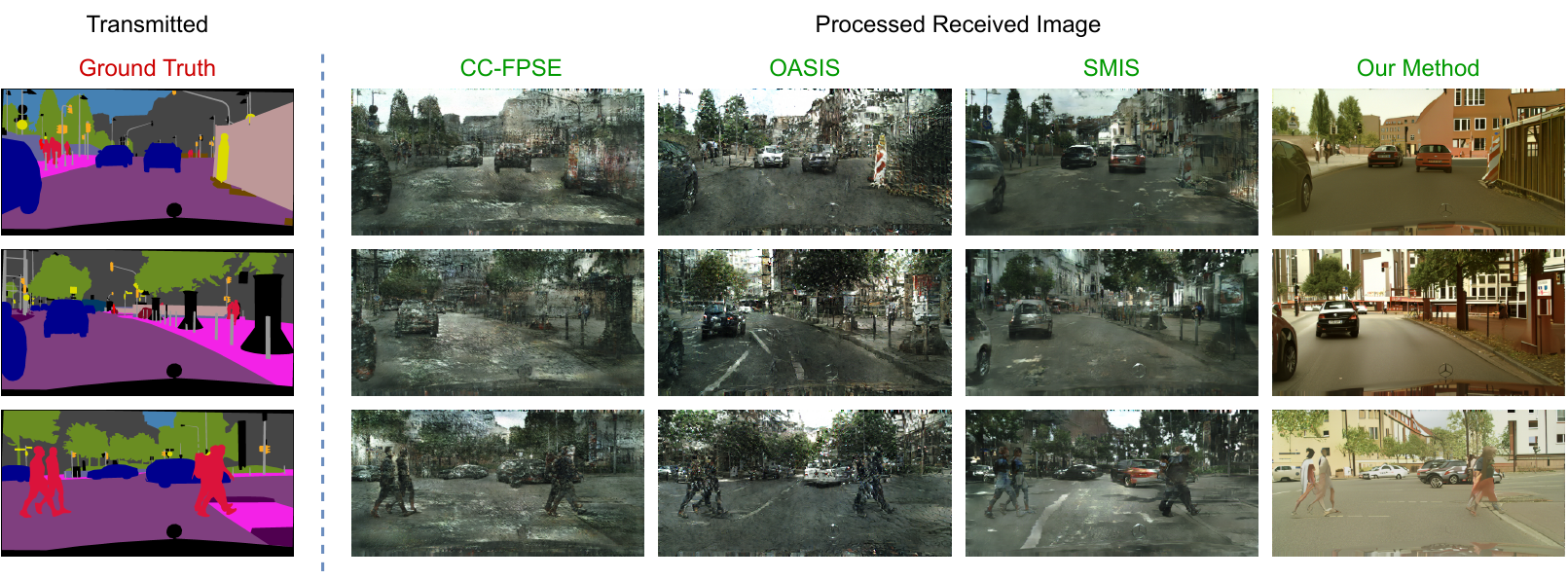}
    \caption{Comparisons among most performing models (CC-FPSE \cite{liu2019learning}, OASIS \cite{schonfeld2021you}, and SMIS \cite{Zhu2020SemanticallyMI}) with $\text{PSNR}=15$. Other methods produce almost noise-only images. Our method produces the best quality samples in which it is easy to recognize objects, cars, and pedestrians, while comparisons generate scenes heavily corrupted by noise.}
    \label{fig:comp1}
\end{figure*}

\subsection{Classifier-free Guidance}
The image quality of conditional diffusion models can be improved through the gradient of the log-probability distribution $\nabla_{\mathbf{x}_t} \log p(\mathbf{y}|\mathbf{x}_t)$ by perturbing the mean with a guidance-scale hyperparameter $s$ \cite{Dhariwal2021BeatGANs}. Whereas previous diffusion models involved a classifier for this procedure \cite{Dhariwal2021BeatGANs}, novel methods directly leverage the generative model to provide the gradient during the sampling step \cite{Ho2021ClassFree}. In our framework, we can disentangle the conditional noise estimation from the unconditional one, by involving the semantic map for the first estimate as $\epsilon_\theta(\mathbf{x}_t|\mathbf{y})$ and the null label for the second one, that is $\epsilon_\theta(\mathbf{x}_t|\mathbf{0})$ \cite{Wang2022SemanticIS}. The gradient of the log-probability distribution is then proportional to the difference between the estimates as

\begin{align}
    \epsilon_\theta(\mathbf{x}_t|\mathbf{y}) - \epsilon_\theta(\mathbf{x}_t|\mathbf{0}) &\propto \nabla_{\mathbf{x}_t} \log p(\mathbf{x}_t|\mathbf{y}) - \nabla_{\mathbf{x}_t} \log p(\mathbf{x}_t)\\
    &\propto \nabla_{\mathbf{x}_t} \log p(\mathbf{y}|\mathbf{x}_t).
\end{align}

Accordingly, the noise estimation is performed through the disentangled component as

\begin{equation}
    \hat{\epsilon}_\theta(\mathbf{x}_t|\mathbf{y}) = \epsilon_\theta(\mathbf{x}_t|\mathbf{y}) + s \cdot (\epsilon_\theta(\mathbf{x}_t|\mathbf{y}) - \epsilon_\theta(\mathbf{x}_t|\mathbf{0})).
\end{equation}

\section{Experimental Evaluation}
\label{sec:exp}

\begin{table*}[]
\caption{Generation quality evaluation of generated images under different channel conditions.}
\label{tab:fid_results}
\centering
\resizebox{\textwidth}{!}{
\begin{tabular}{@{}lccccccc@{}}
\toprule
\multicolumn{1}{l}{Method} & \multicolumn{7}{c}{FID$\times10\downarrow$} \\ \midrule
PSNR       & 100    & 30    & 20     & 15     & 10     & 5 & 1 \\ \midrule
Full image & -      & $\mathbf{6.284}\pm$.053 & 13.684$\pm$.032 & 20.045$\pm$.865 & 28.005$\pm$.878 & 37.931$\pm$.639 & 42.004$\pm$.911   \\
SPADE \cite{park2019SPADE}      & 10.324$\pm$.171 & 14.200$\pm$.179 & 22.971$\pm$.190 & 42.681$\pm$.201 & 55.420$\pm$1.056 & noise & noise   \\
CC-FPSE \cite{liu2019learning}    & 24.590$\pm$.056 & 20.337$\pm$.060 & 26.253$\pm$.049 & 33.166$\pm$.210 & 40.374$\pm$.345 & noise & noise \\
SMIS \cite{Zhu2020SemanticallyMI}       & $\mathbf{8.758}\pm$.162  & 9.147$\pm$.100 & $\mathbf{11.750}\pm$.095 & 14.775$\pm$.129 & 21.373$\pm$.167 & 34.586$\pm$.171 & 44.115$\pm$.412 \\
OASIS \cite{schonfeld2021you}      & 10.403$\pm$.053 & 10.339$\pm$.099 & 16.179$\pm$.122 & 24.892$\pm$.134 & 40.440$\pm$.349 & noise & noise \\
SDM \cite{Wang2022SemanticIS}      &  9.899$\pm$.391 & 16.642$\pm$2.101 & 31.510$\pm$2.926 & noise & noise & noise & noise \\
Our method & 11.848$\pm$.061 & 12.355$\pm$.090 & 14.030$\pm$.201 & $\mathbf{14.008}\pm$.251 & $\mathbf{14.851}\pm$.193 & $\mathbf{15.315}\pm$.349 & $\mathbf{15.989}\pm$.561 \\ \bottomrule
\end{tabular}
}
\end{table*}

\begin{table*}[]
\caption{Perceptual similarity evaluation of generated images under different channel conditions.}
\label{tab:lpips_results}
\centering
\resizebox{\textwidth}{!}{
\begin{tabular}{@{}lccccccc@{}}
\toprule
Method & \multicolumn{7}{c}{LPIPS$\downarrow$} \\ \midrule
PSNR       & 100    & 30    & 20    & 15    & 10    & 5     & 1 \\ \midrule
Full image & - & 0.623$\pm$.074 & 0.684$\pm$.165 & 0.713$\pm$.054 & 0.730$\pm$.156 & 0.747$\pm$.154 & 0.738$\pm$.186 \\
SPADE \cite{park2019SPADE}      & $\mathbf{0.546}\pm$.045 & 0.565$\pm$.072 & 0.603$\pm$.022 & 0.726$\pm$.019 & 0.792$\pm$.115 & 0.824$\pm$.054 & 0.827$\pm$.011 \\
CC-FPSE \cite{liu2019learning}    & $\mathbf{0.546}\pm$.025 & 0.559$\pm$.004 & 0.581$\pm$.009 & 0.620$\pm$.011 & 0.855$\pm$.024 & 0.753$\pm$.032 & 0.812$\pm$.055 \\
SMIS \cite{Zhu2020SemanticallyMI}       & $\mathbf{0.546}\pm$.002 & 0.548$\pm$.030 & 0.561$\pm$.010 & 0.574$\pm$.021 & 0.603$\pm$.027 & 0.649$\pm$.044 & 0.680$\pm$.124 \\
OASIS \cite{schonfeld2021you}      & 0.561$\pm$.032 & 0.564$\pm$.054 & 0.580$\pm$.012 & 0.613$\pm$.073 & 0.679$\pm$.020 & 0.783$\pm$.034 & 0.828$\pm$.122 \\
SDM \cite{Wang2022SemanticIS}      &  0.549$\pm$.061 & 0.543$\pm$.072 & 0.555$\pm$.066 & 0.599$\pm$.043 & 0.606$\pm$.071 & 0.655$\pm$.098 & 0.749$\pm$.119 \\
Our method & 0.590$\pm$.032 & $\mathbf{0.517}\pm$.004 & $\mathbf{0.523}\pm$.011 & $\mathbf{0.542}\pm$.003 & $\mathbf{0.549}\pm$.009 & $\mathbf{0.620}\pm$.023 & $\mathbf{0.609}\pm$.042 \\ \bottomrule
\end{tabular}
}
\end{table*}

\begin{table*}[t]
\caption{Semantic evaluation of generated images under different channel conditions.}
\label{tab:obj_results}
\centering
\resizebox{\textwidth}{!}{
\begin{tabular}{@{}lccccccc@{}}
\toprule
Method & \multicolumn{7}{c}{mIoU$\uparrow$} \\ \midrule
PSNR       & 100    & 30    & 20    & 15    & 10    & 5     & 1 \\ \midrule
Full image &  - & $\mathbf{0.955}\pm$.032 & 0.911$\pm$.155 & 0.906$\pm$.247 & 0.906$\pm$.339 & 0.240$\pm$.193 & 0.110$\pm$.298 \\
SPADE \cite{park2019SPADE}      &  0.909$\pm$.127 & 0.914$\pm$.255 & 0.921$\pm$.315 & 0.812$\pm$.364 & 0.672$\pm$.321 & 0.253$\pm$.288 & 0.313$\pm$.144 \\
CC-FPSE \cite{liu2019learning}    &  0.908$\pm$.045 & 0.908$\pm$.121 & 0.911$\pm$.315 & 0.928$\pm$.345 & 0.852$\pm$.245 & 0.653$\pm$.183 & 0.322$\pm$.284 \\
SMIS \cite{Zhu2020SemanticallyMI}       &  0.909$\pm$.064 & 0.919$\pm$.066    & 0.909$\pm$.214 & 0.931$\pm$.208    & 0.901$\pm$.244 & 0.899$\pm$.290 & 0.876$\pm$.211 \\
OASIS \cite{schonfeld2021you}      &  0.910$\pm$.111 & 0.908$\pm$.191 & 0.912$\pm$.232 & 0.697$\pm$.165 & 0.662$\pm$.356 & 0.345$\pm$.112 & 0.232$\pm$.191 \\
SDM \cite{Wang2022SemanticIS}      &  0.921$\pm$.051 & 0.340$\pm$.022 & 0.333$\pm$.061 & 0.351$\pm$.011 & 0.297$\pm$.021 & 0.256$\pm$.019 & 0.211$\pm$.043 \\
Our method &  $\mathbf{0.940}\pm$.014 & 0.942$\pm$.212 & $\mathbf{0.944}\pm$.297 & $\mathbf{0.945}\pm$.141 & $\mathbf{0.905}\pm$.112 & $\mathbf{0.913}\pm$.214 & $\mathbf{0.925}\pm$.111 \\ \bottomrule
\end{tabular}
}
\end{table*}

\begin{figure*}
    \centering
    \includegraphics[width=\linewidth]{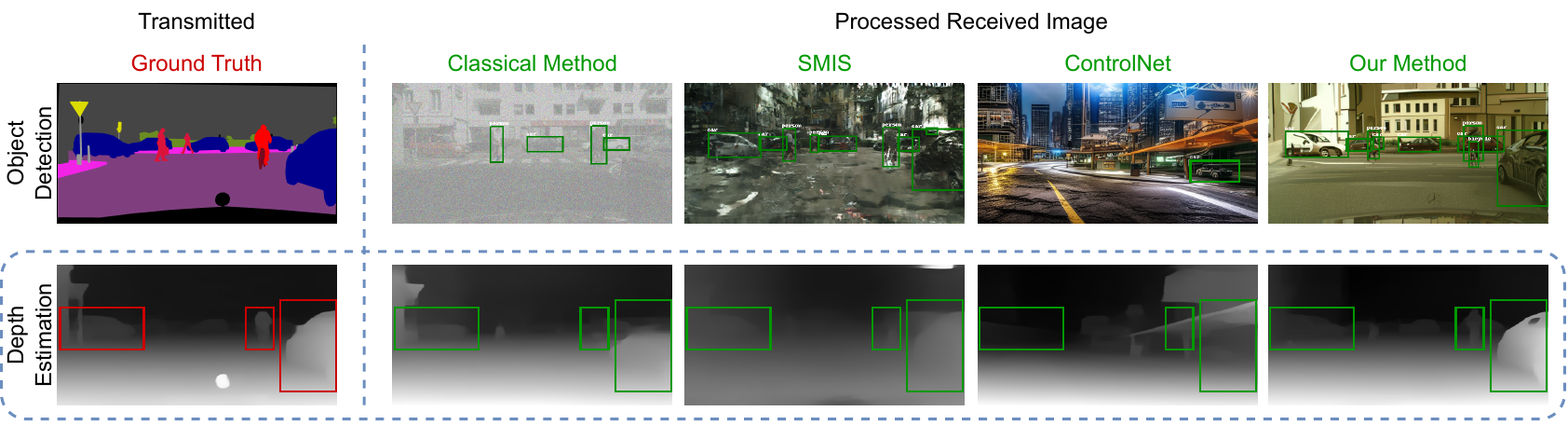}
    \caption{Synthesized images from the transmitted semantics with $\text{PSNR}=10$ for the classical method, SMIS, ControlNet, and our method. The detector can still recognize objects in our generated sample, while other images are too noisy or without preserved semantics such as in ControlNet. The depth estimation confirms the better quality of our generation by correctly estimating distances from objects while producing blurred maps for comparisons.}
    \label{fig:fig1+controlnet}
\end{figure*}

\begin{figure*}[t]
    \centering
    \includegraphics[width=\linewidth]{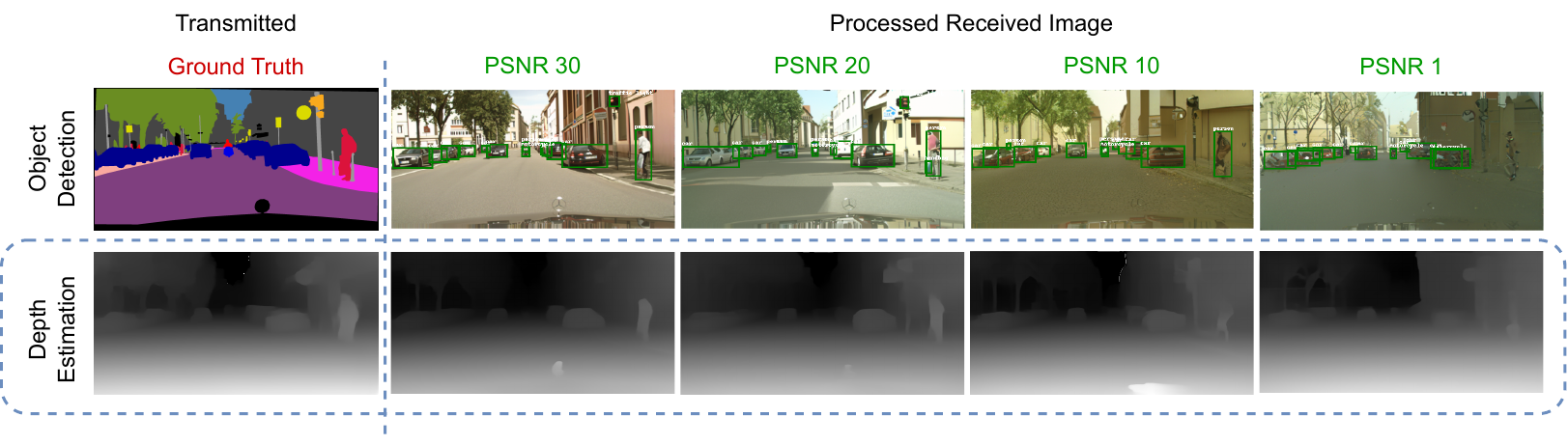}
    \caption{Our method results for different PSNR values of the communication channel. The detector recognizes well cars and pedestrians in all the samples, and the depth estimation is accurate, approaching the ground truth, proving that our method works properly.}
    \label{fig:res1}
\end{figure*}

In this section, we report the experimental setup and the results of the tests. On the receiver side, we regenerate the images from the semantic maps and then perform downstream tasks on them, which consist in object detection, recognition, and depth estimation.


\subsection{Problem Setting}

 In our experiments, we conducted two analyses, each designed to evaluate a different aspect of our system performance: channel robustness and compression efficiency. For the first analysis, we test our system in the presence of distortion of the received semantic map due to propagation over a non-ideal communication channel. In practice, the effect of channel noise on the received segmentation map is rather complicated, as it depends on several factors, such as source and channel coding, symbol constellation, channel model, decoding strategy, and so on. To simplify the analysis and to be consistent with the literature \cite{Gunduz2022GenSem, Shao2021LearningTC}, we modeled this perturbation as additive zero-mean random white noise. However, to check the robustness of our solution, we further test our model under a variety of probabilistic noise models, such as Gaussian, Poisson, and a mixture of noises.

Denoting with $P$ the power of the received segmentation map and with $\sigma^2$ the noise variance, we assess the performance of our method as a function of the 
peak signal-to-noise ratio (PSNR), defined as 

\begin{equation}
    \text{PSNR} = 10 \log \frac{P}{\sigma^2}\, (\text{dB}).
\end{equation}

The second analysis aims at evaluating the semantic distortion/rate tradeoff that the proposed method can achieve. We evaluate this performance in a clean channel scenario.

\textbf{Datasets.} We involve Cityscapes, which contains $35$ classes, and COCO-Stuff, with $183$ classes, as our datasets for training and evaluation. Both datasets comprise instance annotations that we consider in our framework.

\textbf{Training.} We set the guidance scale $s$ equal to $2$ for Cityscapes and $2.5$ for COCO-Stuff, following \cite{Wang2022SemanticIS}. We resize Cityscapes images to $256\times512$, and COCO-Stuff images to $256\times256$. We train the model with PyTorch on a single NVIDIA Tesla V100 GPU (32GB) for the Cityscapes and on a single NVIDIA Quadro RTX8000 (48GB) for COCO-Stuff. During training, for each batch, we sample a certain amount of noise that we apply to the transmitted semantic maps. Specifically, in 30\% of the cases, the maps are transmitted without noise, while in other cases equal probability is assigned to PSNR equal to 30, 25, 20, 15, 10, and 1. In this way, in 70\% of the training batches, the model is conditioned with noisy semantic maps and forced to reconstruct the original clean image. We use a batch size of $4$ in all experiments, a learning rate of $0.0001$ for the AdamW optimizer, and attention blocks at resolutions $32, 16, \text{and }8$ with a number of head channels equal to $64$. The dimension of the features vector in the encoder and in the decoder of the U-Net model is halved at each layer, while comprising a number of channels equal to $[256, 256, 512, 512, 1024, \text{ and } 1024]$. For sampling, we set the number of diffusion steps to $T=1000$ with a linear noise schedule. We used mixed precision for training to reduce the computational complexity. The loss balance term $\lambda$ is set to $0.001$, according to \cite{Wang2022SemanticIS}. Furthermore, we use an exponential moving average of the U-Net network weights with a decay equal to $0.9999$. Figure \ref{fig:blocks} shows the structure of our encoder and decoder blocks with SPADE for semantic conditioning. For compression, we rely on the widespread BPG encoder. The code and the checkpoints are freely available at \url{https://github.com/ispamm/GESCO}.

\subsection{Key Performance Indicators (KPIs)}
The semantic communication framework proposed in this work aims to transmit only the semantic information of an image that is relevant to the application running on top of the information exchange between transmitter and receiver. We considered, as an example, the scenario in which a car transmits images captured by its front-facing camera to a roadside unit (RSU). The task running on the RSU is to identify relevant objects, such as pedestrians, vehicles, and traffic lights, estimate their depth (or distance of these objects from the car) and properly position them relative to each other. This useful information is effectively represented by the semantic segmentation map, whose classes include, among others, the objects mentioned above. Intuitively, the task should be fulfilled in a minimal time, taking into account that communication occurs over a non-ideal wireless channel. Therefore, the proposed method has to compress the information as much as possible while also being robust to channel noise.
To evaluate the performance within this semantic framework, we consider three classes of Key Performance Indicators (KPIs):
1) quality of the regenerated image; 2) ability to recover shapes of relevant objects from the regenerated image; 3) depth estimation, object detection and recognition from the regenerated image.
We describe now the KPIs associated with these classes.

\subsubsection{Quality of the regenerated image}
The quality of the image generated at the receiver side is assessed both qualitatively and quantitatively. We use two well-known metrics commonly employed in computer vision to test the quality of image generative models: the Fréchet Inception Distance (FID) \cite{heusel2017gans} and the Learned Perceptual Image Patch Similarity (LPIPS) \cite{zhang2018unreasonable}:
FID measures the similarity between the distributions of two sets (batches) of real and generated images in a feature space extracted by a deep neural network; LPIPS measures perceptual similarity between two images and is correlated with human visual perception. In both cases, the comparison is carried out in the latent space extracted by the InceptionV3 network \cite{Szegedy2015RethinkingTI} trained for image classification. A low value of FID or LPIPS indicates that the two images are close in the latent space and thus perceived as similar. We do not involve conventional pixel-wise image reconstruction quality metrics, such as PSNR and Structural Similarity Index Measure (SSIM), as we focus on methods able to preserve the semantic content, rather than pixel-level fidelity, as also highlighted in \cite{Agustsson2018GenerativeAN, cicchetti2024MLSP}.

\subsubsection{Recovery of relevant shapes and relative positioning}
To assess the ability to recover the shapes and relative positioning of relevant objects from the regenerated image, we extract the corresponding semantic segmentation map and compare it with the map extracted from the original image using the mean Intersection over Union (mIoU) metric, defined as:
\begin{equation} 
{\rm mIoU} = \frac{1}{n_c} \sum_{i=1}^{n_c} \frac{|\mathbf{s}_i \cap \mathbf{s}'_i|}{|\mathbf{s}_i \cup \mathbf{s}'_i|}, 
\end{equation}
where $n_c$ represents the number of semantic classes, and $\mathbf{s}_i$ and $\mathbf{s}'_i$ are the sets of pixels associated to class $i$ in the original and the regenerated images, respectively.
More specifically, we compute the mIoU metric on the segmentation maps of the generated images obtained through a pretrained model. For this evaluation, we employed DRN-D-105 \cite{Yu2017DilatedRN} on Cityscapes, and MaskFormer \cite{Cheng2021PerPixelCI} on COCO-Stuff. Note that the mIoU evaluation strongly depends on the effectiveness of the pretrained model involved in computing the segmentation maps.

\subsubsection{Depth estimation, object detection and recognition}
In the example of the application we consider, we assume that the task running on the RSU is to estimate the 3D depth of relevant objects from the regenerated 2D images. This is a task that plays a key role in autonomous driving or remote control. In parallel, we run algorithms for the detection and recognition of relevant objects. 
The goodness of the depth estimation is quantified by the Root Mean Square Error (RMSE) metric. To evaluate object detection, we employ the common Mean Average Precision (mAP) with thresholds set at 0.1 for mAP and 0.5 for mAP50. The average precision is defined as the area under the precision-recall curve, and the mAP is the mean of this metric over all object classes.
As downstream methods for these tasks, we employ two well-known methods in the literature: We use DEtection TRansformer (DETR) \cite{Carion2020DETR} for object detection and Dense Prediction Transformer (DPT) for depth estimation \cite{Ranftl2021VisionTF}, \cite{fonder2021m4depth}.


\subsection{Comparison with alternative methods}

We compare our proposal with a classical communication method that directly transmits the image and with methods that exploit generative models. We consider well-known semantic image synthesis models such as SPADE \cite{park2019SPADE}, CC-FPSE \cite{liu2019learning}, SMIS \cite{Zhu2020SemanticallyMI}, OASIS \cite{schonfeld2021you}, SDM \cite{Wang2022SemanticIS}, and ControlNet \cite{Zhang2023AddingCC}. Notably, SPADE, CC-FPSE, SMIS, and OASIS are based on generative adversarial networks (GANs), while SDM and ControlNet are based on diffusion models. None of them has a denoising component. We consider different channel scenarios, ranging from extremely degraded conditions to perfect transmissions.\\

\begin{table}[t]
\caption{Semantic, perceptual similarity, and generation quality evaluation with fixed $\text{PSNR}=10$ on the COCO-Stuff dataset.}
\label{tab:coco}
\centering
\scriptsize
\begin{tabular}{@{}lccc@{}}
\toprule
Method & mIoU$\uparrow$ & LPIPS$\downarrow$ & FID$\times10\downarrow$  \\ \midrule
Full image & 0.331$\pm$.145 & 0.687$\pm$.003 & 40.562$\pm$2.513 \\
SPADE \cite{park2019SPADE}      & noise & noise & noise \\
CC-FPSE \cite{liu2019learning}    & noise & noise & noise \\
SDM \cite{Wang2022SemanticIS}      & noise & noise & noise \\
Our method & $\mathbf{0.365}\pm$.096 & $\mathbf{0.683}\pm$.011 & $\mathbf{36.664}\pm$1.527 \\ \bottomrule
\end{tabular}
\end{table}

\begin{table}
\centering
\caption{Quantitative metrics for downstream tasks (object detection and depth estimation).}
\label{tab:downstream_metrics}
\begin{tabular}{@{}lccc@{}}
\toprule
Task       & \multicolumn{2}{c}{Obj det.} & Depth est. \\ \midrule
Model      & mAP$\uparrow$           & mAP50$\uparrow$        & RMSE$\downarrow$       \\ \midrule
Semantic Map       & -         & -        &   208.984         \\
ControlNet \cite{Zhang2023AddingCC}      & 0.018         & 0.061        &  199.211          \\
SMIS \cite{Zhu2020SemanticallyMI}      & 0.230         & 0.451        &  44.102          \\
Ours       & $\mathbf{0.390}$         & $\mathbf{0.666}$        &   $\mathbf{14.530}$         \\ \bottomrule
\end{tabular}
\end{table}

\textbf{Quality of regenerated images.}
Figures~\ref{fig:comp} and \ref{fig:comp1} show a comparison between our method and a set of competitive methods. More specifically, the images in the leftmost column of Fig.~\ref{fig:comp} show two examples of semantic maps. These images are sent through a non-ideal channel that adds white noise with PSNR = 10 dB. To have an idea of the impact of such a noise on the true image, the immediately adjacent column shows the images that would be observed with such an additive noise. The remaining columns show the images regenerated from the received noisy map using alternative methods. More specifically, in Fig.~\ref{fig:comp} we compare our method with SPADE and SDM methods, with PSNR = 10 dB,  while in Fig.~\ref{fig:comp1} 
we compare it with CC-FPSE, OASIS, and SMIS methods, assuming PSNR=15 dB.
From both figures, we can see that our method is more robust than the others. In all these examples, we can see that all objects deemed as relevant in the image, such as pedestrians and cars, for example, are well regenerated by our method. Interestingly, in the case of noisy conditions, diffusion models may sometimes generate images lacking details. 
Nevertheless, since we only transmit one-hot encoded semantic maps, we are able to make the transmission much more robust against additive noise than other methods. Of course, our regenerated objects are different from the original ones, but they preserve shape and relative positioning very well. Hence, in all applications where it is not really required to reconstruct the original images, but only to reconstruct an overall scene, with correct shapes and positions of all relevant objects, our method shows clear advantages with respect to alternative ones.

The previous examples provide only a qualitative comparison over a limited set of images. To provide a quantitative assessment, in Tables~\ref{tab:fid_results}, \ref{tab:lpips_results}, and \ref{tab:obj_results}, we report the FID, LPIPS, and mIoU scores obtained for alternative methods, under different values of PSNR. From these tables, we can see that our method outperforms all others, in terms of FID, when the PSNR falls below 15 dB, while, in terms of LPIPS, our method performs the best as soon as the PSNR falls below 30 dB. Similarly, also the ability to recover the correct shapes from the noisy map, quantified by the mIoU metrics, is the best over almost all PSNR values, as shown in Table~\ref{tab:obj_results}. Furthermore, in Table~\ref{tab:coco} we report the values of mIoU, LPIPS, and FID, for alternative methods running on the  COCO dataset, in the presence of noise (PSNR = 10 dB). We can see that the alternative approaches, at this level of noise, produce only noisy samples, while our method provides results not too far from those obtainable from the original image.\\

\textbf{Depth estimation.}
In this section, we evaluate the performance of the downstream task of recovering the depth of relevant objects from the received noisy map. In principle, if the task is only depth estimation, we could think of applying depth estimation methods directly to the received map. The fundamental question then is whether it is necessary to regenerate images at the receiver side or not. To answer this question, in Table \ref{tab:downstream_metrics} we report the RMSE of the depth estimation performed by DPT on the semantic map and on the images regenerated by SMIS and by our method.  We can clearly see that DPT performs very poorly on the semantic map. This happens because DPT is trained on realistic images and then it does not generalize well when applied to semantic maps. Furthermore, when transmitting the entire map over the channel, the map can be severely degraded by the channel noise. This behavior motivates the need to properly regenerate the image from our one-hot-encoded maps before using DPT or any other downstream model to perform depth estimation or additional downstream applications. 
Figure~\ref{fig:fig1+controlnet} reports an example of a semantic map, together with the result of applying the detection algorithm DETR (top row) and the depth estimator DPT (bottom row) to a set of images regenerated using alternative methods (SMIS, ControlNet and ours). The depth estimation is represented by the gray level of each pixel: The lighter pixels are the closest ones, while the darker ones are the farthest. In all cases, the received semantic map is corrupted by noise, with PSNR = 10 dB. We also report (leftmost column) the ground truth, represented by the semantic map (top) and the depth estimation map (bottom) extracted directly from the original image, as a comparison term. In the second column, from the left, we also report the original image, corrupted by the same level of noise and the corresponding depth estimation. From Fig.~\ref{fig:fig1+controlnet}, we can see that the depth recovered from our method is much closer to the original one than the alternative methods. It is also interesting to observe that, in the original image, there is a building in the left part of the image, which is not captured by the semantic map because ``building'' is not a relevant class, but it is captured by the depth estimation map. As a consequence, since our method is based solely on the semantic map, it is unable to visualize the building on the left side or estimate its depth, while it is perfectly capable of visualizing crucial objects, such as the nearby car on the right side of the image.
We also evaluated the robustness of the method to additive noise. To this end, Figure~\ref{fig:res1} shows the images generated from maps corrupted by different levels of noise, together with the object detection (green boxes in the top row) and estimated depth (bottom row). We can notice that, even in the case of very low PSNR, DETR is still able to recognize most of the relevant objects. Furthermore, also the depth estimation gives consistent results across very different conditions. 

To perform a quantitative comparison, in terms of object detection and depth estimation, in Tab.~\ref{tab:downstream_metrics} we report
the mAP and mAP50 metrics for object detection and the RMSE for depth estimation, in the case where PSNR = 10 dB. We compare our method with 
SMIS \cite{Zhu2020SemanticallyMI} and ControlNet methods
\cite{Zhang2023AddingCC}, a recent powerful conditional method for diffusion models.
From Tab.~\ref{tab:downstream_metrics}, we can see that the best performing alternative to our method is SMIS and that our method outperforms SMIS across all metrics. We can also observe that the estimated depth from the solely transmitted semantic map has a very large RMSE, thus confirming the need for regenerating realistic images at the receiver side before running DPT. From Tab.~\ref{tab:downstream_metrics}, we also notice that  ControlNet has a poor performance compared to SMIS and to our method. This happens because the iterated conditioning occurring in the diffusion model is very sensitive to the quality of the conditioning map. If the map is corrupted, the iterated conditioning can lead to a very poor result. Conversely, being based on a GAN method, SMIS does not suffer from this iterated conditioning. Our method is also more robust than ControlNet because, even though we also use a diffusion model, we can recover a semantic map of much higher quality at the receiver side, under the same transmission rate, because we only transmit the semantic map and we also apply our algorithm to reduce the impact of noise on the reconstruction of the semantic map.\\

\begin{figure}
    \centering
    \includegraphics[width=\linewidth]{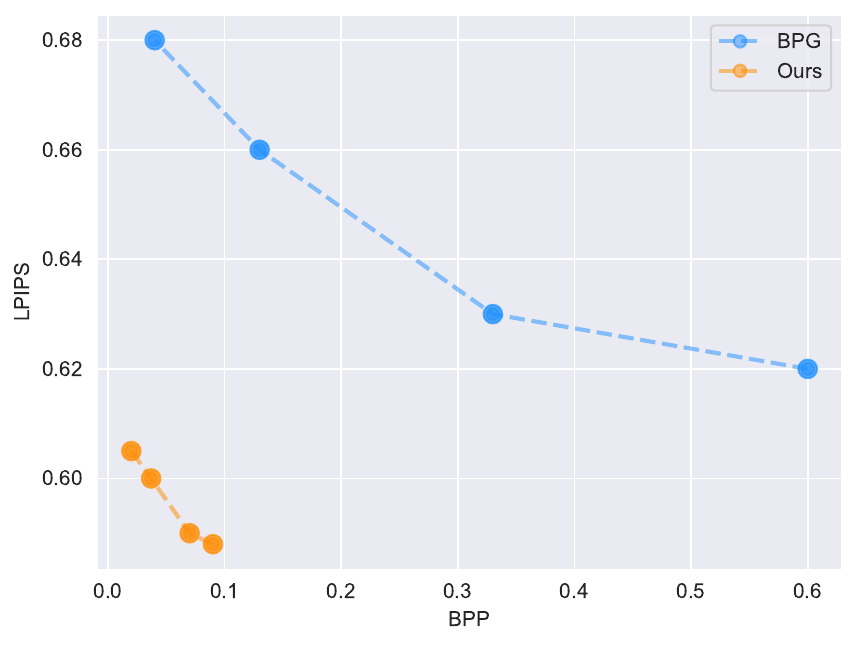}
    \caption{Perceptual similarity (LPIPS) of images under low bit per pixel (BPP). The proposed method crucially reduces the bit rate while obtaining better LPIPS scores.}
    \label{fig:lpips_bpp}
\end{figure}

\begin{figure}
    \centering
    \includegraphics[width=\linewidth]{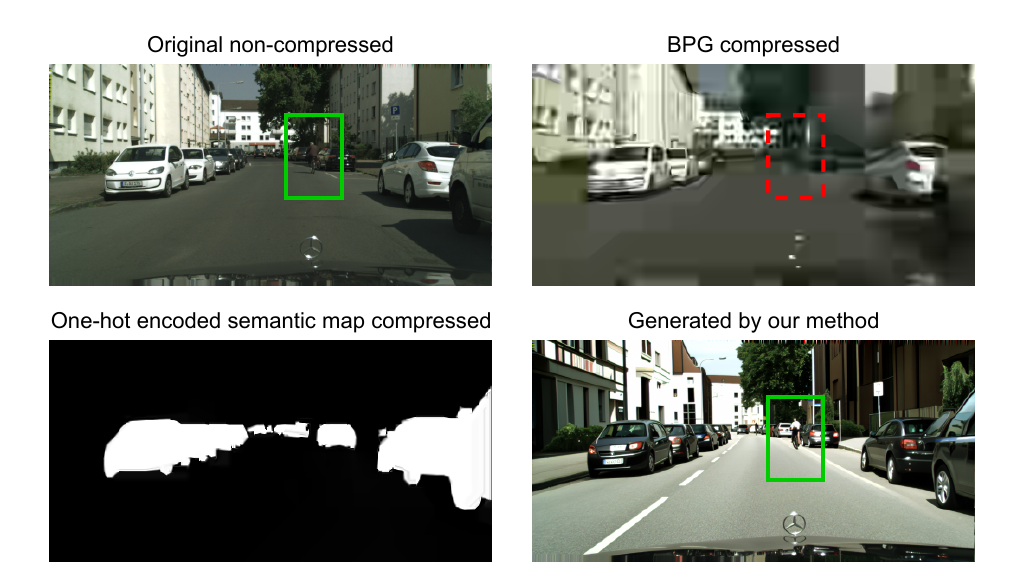}
    \caption{Example of how the BPG compression (quantizer parameter = $0.51$, full image with BPP = $0.09$ and semantic map with BPP = $0.005$) affects the original image and a sample of the one-hot encoded maps. The compressed image loses informative content, while the one-hot encoded maps are minimally affected by the compression, and therefore, the proposed method is able to regenerate an image full of original semantic content. As an example, the biker disappears on the compressed image, while it is still easily recognizable in the regenerated one.}
    \label{fig:compress}
\end{figure}

\begin{figure}
    \centering
    \includegraphics[width=\linewidth]{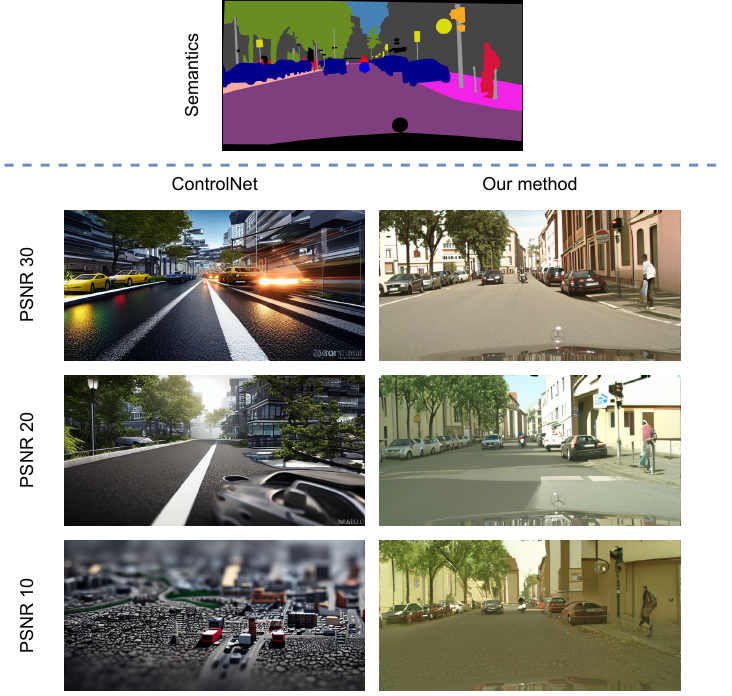}
    \caption{Generated samples under different channel conditions (PSNR in $\{30, 20, 10\}$) by ControlNet and by the proposed method.}
    \label{fig:suppl1_controlnet}
\end{figure}

\textbf{Semantic distortion/rate tradeoff.} A fundamental aspect of communication is the trade-off between the complexity of a representation (expressed in terms of number of bits) and the distortion between the original multimedia content and such a representation. In a semantic framework, it is important to generalize this principle using a definition of distortion associated with the semantics of the content. In our case, the semantic distortion can be measured, for example, by the LPIPS metric, while the complexity is measured by the number of bits-per-pixel (BPP) obtained encoding the image to be transmitted, either the original one or the semantic map, using the well-known image coding Better Portable Graphics (BPG) \cite{albalawi2015hardware}. To this purpose, in Fig.~\ref{fig:lpips_bpp} we report the LPIPS value vs. the BPP value obtained by applying BPG to the original image and to the one-hot-encoded segmentation maps. The evaluation has been carried out on the Cityscapes dataset with images resized to $256\times512$, compressing both the full original image and the one-hot-encoded semantic maps. From Fig.~\ref{fig:lpips_bpp}, we can clearly see a striking advantage obtained with our method. A visual explanation for these advantages can be seen in Fig.~\ref{fig:compress}, where we show the original image, the image reconstructed from the BPG-compressed one, a sample of the compressed one-hot-encoded map, and the corresponding regenerated image. Differently from BPG, which can significantly impair the detection of distant, and therefore small, objects in heavily compressed images, the quality of the reconstructed map using our method is scarcely affected because the map is constant within the patch corresponding to each relevant object. This leads to a high-quality regenerated image, with crucial objects like the cyclist still recognizable, contrary to the compressed original image. Being the one-hot-encoded semantic maps originally encoded with 1 bit as they are binary maps (0-1), they can reach much lower BPP with respect to compressing original images while barely affecting their quality, as shown in the second row of Fig.~\ref{fig:compress}, where the compressed map does not show corruptions or degradations. It is worth pointing out that the results shown above have all been obtained using the highest compression ratio achievable with BPG when applied to the image (0.09 BPP) and to its semantic map (0.005 BPP).
\textbf{Need for Pretrained Generative Models?} Recently, a common way of working is taking large pretrained models and applying them in downstream tasks of interest. Usually, the proprietary companies of such models do not release either huge datasets,  checkpoints, or model specifications. In a communication scenario, although being an easy and straightforward solution as increasing the dataset yields better learning ability, adopting large pretrained models may incur some issues. These models have not been engineered and trained for such a real-world problem. Thus, they are not robust to heavy corruption due to communication over noisy channels. This may result in noisy, imprecise, and corrupted generated content that cannot be considered reliable from a communication perspective.
To provide some experimental validation of our approach, we compare our method with ControlNet \cite{Zhang2023AddingCC}. ControlNet is mounted on top of the state-of-the-art generative model Stable Diffusion \cite{Rombach2021HighResolutionIS}, which has been trained on the LAION-5B dataset \cite{schuhmann2022laionb} comprising $5.8 \times 10^9$ images. Conversely, our dataset is composed of $3 \times 10^3$ images. Compared to the training of Stable Diffusion on the LAION-5B dataset, our training requires lower computational resources and time, making it affordable with a constrained budget.
We report some examples in Fig.~\ref{fig:suppl1_controlnet} where the top image reports the ideal semantic map and the two columns show the images regenerated by ControlNet (left) and our method (right), under different values of PSNR. This figure shows that, especially when the noise is such that the PSNR falls below 20 dB, our generative model is much more robust, in spite of having been trained on a dataset six orders of magnitude smaller than the dataset used by ControlNet. Indeed, while for good channel conditions, i.e., PSNR$=30$, ControlNet generates a meaningful sample, for lower PSNR values, it may completely lose its ability to preserve the semantics of the generated samples, making it unusable in communication systems affected by noise. Conversely, our method is more suitable for task-oriented communications than large-scale pretrained generative models.\\

\begin{table}
\centering
\caption{Results for different channel noises with PSNR=10.}
\label{tab:noises}
\begin{tabular}{@{}l|lcc@{}}
\toprule
Noise & Model & \multicolumn{1}{c}{LPIPS$\downarrow$} & \multicolumn{1}{c}{mIoU$\uparrow$} \\ \midrule
\multirow{2}{*}{AWGN} & SMIS \cite{Zhu2020SemanticallyMI} & 0.603 & 0.901 \\
                      & Ours & $\mathbf{0.549}$ & $\mathbf{0.905}$ \\ \midrule
\multirow{2}{*}{Poisson} & SMIS \cite{Zhu2020SemanticallyMI} & 0.631 & 0.795\\
                         & Ours & $\mathbf{0.595}$ & $\mathbf{0.842}$ \\ \midrule
\multirow{2}{*}{Mixture} & SMIS \cite{Zhu2020SemanticallyMI} & 0.639 & 0.822 \\
     & Ours & $\mathbf{0.599}$ & $\mathbf{0.864}$ \\ \bottomrule
\end{tabular}
\end{table}

\textbf{Tests with Different Channel Noises.} In the communication over a wireless channel, the distortion of the image caused by imperfect channel conditions can be very complicated to handle. In all previous examples, we modeled this distortion as an additive white Gaussian noise (AWGN). In this paragraph, we report the results of further experimentation, where the model is trained using AWGN, but the test is performed using a different kind of noise, i.e. Poisson noise and a mixture of Poisson and Gaussian noises. In Tab.~\ref{tab:noises} We report the results for these different channel noises, with PSNR = 10, obtained by the SMIS method and by our method. We can clearly see a higher robustness of our method vs. SMIS, over all noise scenarios considered in the experiment. As future work, further distortions may be considered, such as screen distortions \cite{Fu2024WaveRecoverySW, palle1}.

\begin{table}[]
\centering
\caption{LPIPS scores for Rayleigh fading channel and trace-driven evaluation with DeepMIMO.}
\label{tab:trace}
\begin{tabular}{lccc|cccc}
\toprule
Channel    & \multicolumn{3}{c|}{Rayleigh} & \multicolumn{4}{c}{DeepMIMO}   \\ \midrule
PSNR       & 30       & 20       & 15      & 100 & 10 & 5                     & 1 \\ \midrule
Our method &  0.56        &  0.58        &   0.62      & 0.65 & 0.66  & 0.65 & 0.64  \\ \bottomrule
\end{tabular}
\end{table}

\textbf{Trace-driven Evaluation.}
To assess the reliability of our approach under realistic propagation conditions, we further evaluate the framework using trace-driven wireless channels derived from the DeepMIMO v3 dataset~\cite{Alkhateeb2019}. Specifically, we use scenario \texttt{O1\_60}, which models a 60 GHz urban canyon environment based on ray-tracing calibrated to real-world measurements. Each semantic map is split in blocks that are multiplexed using a 256-subcarrier OFDM waveform, each subcarrier carrying a QPSK symbol. The channel is frequency-selective and the frequency response is randomly drawn from the DeepMIMO trace. At the receiver side, white Gaussian noise is added to the received signal, with an assigned signal-to-noise ratio, and zero-forcing equalization is applied to each subcarrier.
This setup provides a realistic yet reproducible approximation of field conditions, capturing key effects such as multipath delay spread, spatial correlation, and frequency-selective fading. Unlike purely synthetic fading models, these channel responses reflect practical deployment scenarios encountered in mmWave communication systems.
As shown in Table~\ref{tab:trace}, our framework maintains high performance under this realistic setting, barely showing any loss in performance even in the case of severe degradations. This is on par with the results obtained under AWGN and the additional Rayleigh fading (reported as well in Table \ref{tab:trace}), and confirms the robustness and generalization of the proposed semantic generative pipeline to real-world wireless impairments.




\begin{figure*}
    \centering
    \includegraphics[width=0.9\linewidth]{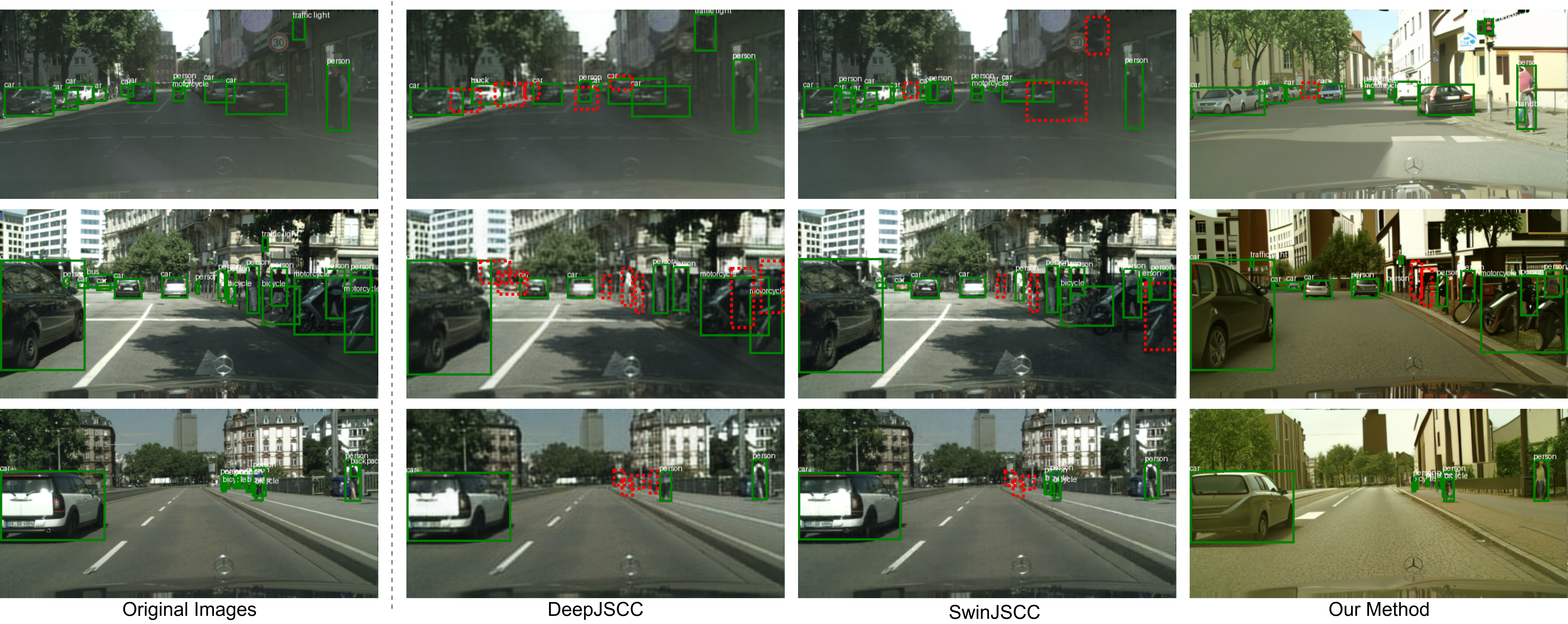}
    \caption{Comparison with DeepJSCC and SwinJSCC. Green boxes are objects detected by DETR, red boxes are objects that the object detector cannot recognize due to bad reconstruction by DeepJSCC and SwinJSCC.}
    \label{fig:djscc}
\end{figure*}

\begin{figure}
    \centering
    \includegraphics[width=\linewidth]{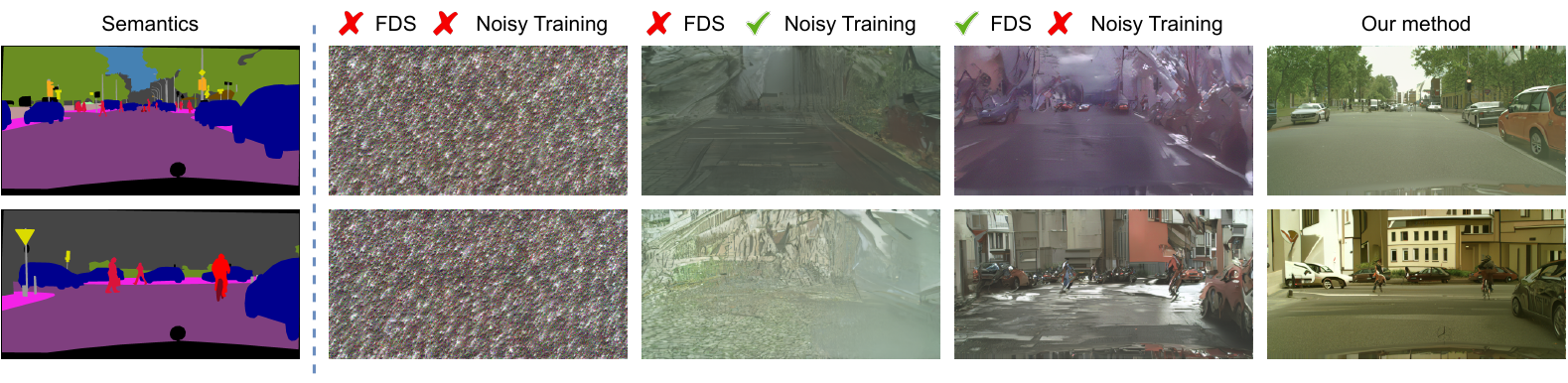}
    \caption{Generated samples from ablation studies with $\text{PSNR}=10$. Samples without FDS and noisy training are clearly noisy. Then, both FDS and noisy training help improve sample quality.}
    \label{fig:abl1}
\end{figure}

\textbf{Comparison with JSCC-related Methods.} Even though our method does not involve joint source-channel coding (JSCC), since JSCC is receiving a large interest in the scientific community, in this section, we provide an analysis of the trade-off between semantic fidelity, bandwidth usage, and computational complexity, comparing our method with DeepJSCC \cite{Kura2020Deepjscc} and SwinJSCC \cite{Yang2025Swinjscc}. Figure \ref{fig:djscc} compares the outputs of DeepJSCC, SwinJSCC, and our method. The comparison is carried out by checking the capability to recognize relevant objects in the images reconstructed by different methods using the state-of-the-art DEtection TRansformer (DETR) method \cite{carion2020end}. In particular, the green boxes identify the correct objects position, whereas the red boxes identify the missed objects. We can see that, while DeepJSCC methods are able to retain the overall scene structure and color information, they fail to recover small or distant objects, such as pedestrians or cars, which are indeed the critical objects for downstream understanding. 
In contrast, our method, even if not able to reproduce a pixel-wise reconstruction of the overall image, is much better in preserving the shape and positioning of the relevant objects from the regenerated images. Quantitatively, in the object detection task on reconstructed images, DeepJSCC obtains a mAP of 0.232, SwinJSCC of 0.315, while our method outperforms both by achieving a mAP of 0.390. Additionally, the proposed system operates at a much lower bit-per-pixel (BPP) rate, transmitting only 0.02 BPP versus 0.17 BPP for DeepJSCC, highlighting a significantly more efficient use of bandwidth.
From a computational perspective, we measure inference times on an NVIDIA RTX 4080 GPU with 16GB of memory. DeepJSCC requires approximately 8 seconds per image. By contrast, each denoising step of our diffusion model takes approximately 0.18 seconds. Clearly, the multi-step nature of diffusion models introduces latency, and this is an aspect that deserves further investigation. It is useful to mention that recent research on accelerated sampling and model quantization offers promising directions to reduce inference time without sacrificing semantic fidelity.

\begin{table}[t]
\caption{Ablation results on the Cityscapes dataset.}
\label{tab:abl}
\centering
\scriptsize
\begin{tabular}{cccc}
\toprule
FDS & Noisy training & LPIPS & mIoU \\ \midrule
\color{red} \ding{55} & \color{red} \ding{55} & noise & noise  \\
\color{red} \ding{55} & \color{green} \ding{51} & 0.665 & 0.613 \\ 
\color{green} \ding{51} & \color{red} \ding{55} & 0.663 & 0.713 \\ 
\color{green} \ding{51} & \color{green} \ding{51} & $\mathbf{0.549}$ & $\mathbf{0.905}$ \\ \bottomrule
\end{tabular}
\end{table}

\begin{table}
\centering
\caption{Ablation study with fixed channel PSNR=10 for the proposed FDS block against a Swin-UNet (SUNet) denoising network \cite{fan2022sunet}.}
\label{tab:fds_ablation}
\resizebox{\linewidth}{!}{%
\begin{tabular}{@{}lccccc@{}}
\toprule
Method & Params & FLOPS & Memory & LPIPS$\downarrow$ & mIoU$\uparrow$ \\ \midrule
SUNet \cite{fan2022sunet}  & 99M    & 60G ($+1400\%$)   &  1.1GB              & 0.575 & 0.869 \\ 
FDS (ours)    & 0M     &  4G     & 0.0GB           & $\mathbf{0.549}$ & $\mathbf{0.905}$     \\
\bottomrule
\end{tabular}}
\end{table}



\subsection{Ablation Studies}
In this subsection, we perform ablation tests to corroborate our methodological choices. First, we study the inference performance with and without the proposed FDS block and without the noisy maps during training, fixing the PSNR value to $10$. Table~\ref{tab:abl} shows the effectiveness and the importance of both the proposed noisy training and the FDS module in the inference phase. Figure~\ref{fig:abl1} allows for a visual inspection of the generated results without the proposed methods. While the semantic diffusion model alone (\textcolor{red}{\ding{55}} FDS \textcolor{red}{\ding{55}} Noisy Training) produces only noise, using both FDS and noisy training effectively helps improve performance.


Second, to further evaluate the effectiveness of the proposed fast denoising semantic (FDS) block, we compare it with a Swin UNet Transformer (SUNet) \cite{fan2022sunet} for image denoising. Our proposed method has several advantages over a denoising network. First, it has no trainable parameters and then it does not require retraining when the scenario changes. Second, it has very light computations, therefore it does not affect the number of FLOPs of the model or the memory for checkpoint storage, as instead required by a denoising network. Table~\ref{tab:fds_ablation} shows the results of the proposed FDS module against the SUNet denoising network. The table confirms our intuition and the denoising model adds a consistent number of FLOPs to the computations, as well as more storage memory for saving the checkpoints to obtain similar results, actually worse than the FDS module.

\section{Conclusion}
\label{sec:conc}
This paper presented a novel generative semantic communication framework whose core blocks include the extraction of the semantic segmentation map, its coding, and the generation of images at the receiver side, conditioned on the semantic map using the proposed semantic diffusion model. 
We made the whole framework robust to bad channel conditions by training the semantic diffusion model with noisy semantic maps and by inserting a fast denoising semantic block to improve the quality of the inferred image. 
We have shown that the proposed method enables a significant reduction of the number of bits to be transmitted necessary to achieve, at the receiver side, a given semantic distortion, measured by a perceptual image quality parameter, such as LPIPS.  Our performance evaluation highlights that the proposed framework generates semantically consistent samples even in the case of highly degraded channel conditions, outperforming all other competitors. 
This empirical robustness is consistent with the ELBO-based variational guarantee of \ref{subsec:losses}, which ensures convergence of the learned reverse process to the true conditional distribution as model capacity and diffusion steps grow. As future work, we plan to mitigate the long inference times and high computational requirements, which are the great challenges of integrating diffusion models in communication systems, through strategies such as quantization \cite{pignata2024icassp, Liu2024DiffusionbasedGM}.




\bibliographystyle{ieeetr}
\bibliography{DiffSC}

\section*{Acknowledgment}
The authors would like to acknowledge the PhD student Luigi Sigillo for the insightful discussions and suggestions and the PhD student Redemptor Jr Laceda Taloma for the help with some technical issues.

\end{document}